\documentclass[journal]{IEEEtran}
\usepackage{algorithm}
\usepackage{algorithmicx}
\usepackage{algpseudocode}
\usepackage{amsmath}
\usepackage{indentfirst}
\usepackage{amsmath}
\usepackage{graphicx}
\usepackage{subfigure}
\usepackage{bm}
\usepackage{amssymb}
\usepackage{multirow}
\usepackage[colorlinks,
            linkcolor=red,
            anchorcolor=black,
            citecolor=green]{hyperref}
\usepackage{amssymb}
\usepackage{booktabs}
\usepackage{float}
\ifCLASSINFOpdf
\else
\fi
\everymath{\displaystyle}
\begin{document}
\title{Learning Physical Dynamics for Object-centric Visual Prediction}

\author{Huilin Xu,~\IEEEmembership{Student Member,~IEEE,}
Tao Chen,~\IEEEmembership{Senior Member,~IEEE,}
        Feng Xu,~\IEEEmembership{Senior Member,~IEEE,}}
        % <-this % stops a space
% \thanks{H. Xu is pursuing the Ph.D with the Key Laboratory of EMW information, Fudan University, China (e-mail: hlxu21@m.fudan.edu.cn)}% <-this % stops a space
% \thanks{F. Xu is with the Key Laboratory of EMW information, Fudan University, China (e-mail: xf@fudan.edu.cn)}% <-this % stops a space
% % \thanks{Manuscript received April 19, 2005; revised August 26, 2015.}


\markboth{Journal of \LaTeX\ Class Files,~Vol.~14, No.~8, August~2015}%
{Shell \MakeLowercase{\textit{et al.}}: Bare Demo of IEEEtran.cls for IEEE Journals}



\maketitle

\begin{abstract}
The ability to model the underlying dynamics of visual scenes and reason about the future is central to human intelligence. Many attempts have been made to empower intelligent systems with such physical understanding and prediction abilities. However, most existing methods focus on pixel-to-pixel prediction, which suffers from heavy computational costs while lacking a deep understanding of the physical dynamics behind videos. Recently, object-centric prediction methods have emerged and attracted increasing interest. Inspired by it, this paper proposes an unsupervised object-centric prediction model that makes future predictions by learning visual dynamics between objects. Our model consists of two modules, perceptual, and dynamic module. The perceptual module is utilized to decompose images into several objects and synthesize images with a set of object-centric representations. The dynamic module fuses contextual information, takes environment-object and object-object interaction into account, and predicts the future trajectory of objects. Extensive experiments are conducted to validate the effectiveness of the proposed method. Both quantitative and qualitative experimental results demonstrate that our model generates higher visual quality and more physically reliable predictions compared to the state-of-the-art methods.
\end{abstract}


% \fi
\IEEEpeerreviewmaketitle

\section{Introduction}

\IEEEPARstart{U}{nderstanding} and predicting physical phenomena in the real world is the core ability of human intelligence and plays a vital role in everyday life. It assists people in avoiding potential risks and making decisions. Players in a table tennis match can infer the trajectory of the ping-pong ball after it collides with the table by observing the ball's early movement and responding correctly. This requires knowledge of the underlying physical laws of the world. Thus, developing the ability to understand, model, and predict physical processes through vision is a key step toward future general artificial intelligence. How to learn the physical dynamics behind visual observations and reason about the future is a fundamental problem that shall be studied and has a wide range of potential applications including automatic driving \cite{ding2022egospeed}, robotics \cite{zang2022robot} and model predictive control \cite{ye2020object}.

Over the past few years, researchers have made a number of attempts \cite{duan2022survey}, \cite{weihs2022benchmarking}. Various tasks and benchmark datasets have been established to assess the physical reasoning capabilities of intelligent systems \cite{bakhtin2019phyre}, \cite{riochet2020intphys}, \cite{yi2020clevrer}. While previous work focuses on inferring object's physical properties \cite{wu2016physics}, \cite{kandukuri2022physical} or event predictions \cite{lerer2016learning}, \cite{bear12021physion}, this paper focuses more on how an agent can learn complex dynamics from single or multiple visual images and predict future trajectories or video frames, which we refer to as visual dynamics and visual prediction tasks.

\begin{figure}[ht]  %允许各个位置
 \center{        
 \includegraphics[scale=0.46]{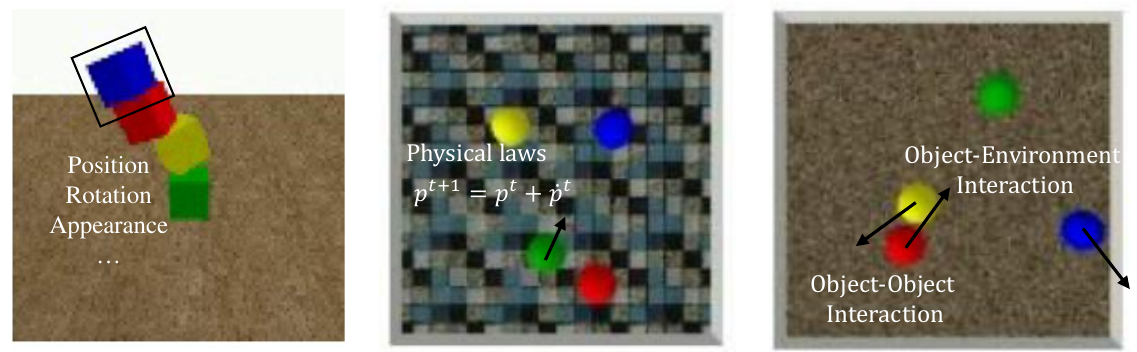}}  %缩放比例的倍数
 \caption{The motivation of the proposed model on visual prediction task. Left: an object representation with rich expressive power is beneficial for prediction. Middle: incorporating prior physical knowledge help simplify complex dynamics and boost model's performance. Right: context information, such as object-object interaction and object-environment interaction, should be considered when inferring future object states.
}
 \label{motivation}
\end{figure}

Conventionally, visual prediction task is treated as an image translation problem, using seq2seq structure to generate future frames in high-dimensional image space \cite{oprea2020review}, \cite{wang2022predrnn}. However, such approaches often have heavy computational costs while lacking insight into physical interaction mechanisms between objects. Recently, an object-centric prediction paradigm has emerged. Unlike dense pixel prediction, this class of approaches performs object-level prediction in low-dimensional state space and produces future trajectories of each object. The prediction results are converted back into future images via a render engine.

Object-centric prediction methods are classified into two categories depending on whether object annotation is required: supervised and unsupervised methods.  Supervised methods bypass the challenge of extracting object-centric representations from images, thus facilitating the subsequent prediction process. However, they rely heavily on time-consuming extensive annotations or additional information such as depth maps, semantic maps \cite{wu2017learning}, \cite{janner2019reasoning}, etc. On the contrary, unsupervised work has attempted to obtain object states from motion cues contained in consecutive frames without the need for annotation \cite{zhu2018object}, \cite{schmeckpeper2021object}. However, existing unsupervised object-centric approaches still suffer from the following issues:
\begin{itemize}
\item [(1)] Trade-off between physical interpretability and expressiveness of object representation. Simple coordinates have explicit physical properties but lose many details, such as appearance. The 1D feature vectors or 3D feature maps extracted by convolutional neural networks (CNNs) contain rich information, but lack physical meaning and increase computational burden. 
\item [(2)] Weak physical compatibility. Object representation, which has poor physical interpretability, hinders the integration of the predictive model with existing well-known physical laws, such as Newton's laws of motion. The weak physical compatibility makes it difficult for these models to be benefited from prior physical knowledge. 
\item [(3)] Contextual information is usually ignored while it plays an essential role in inferring the future, especially in complex environments. The interaction of the object with its surroundings and other objects, as well as the temporal evolution of its own state (velocity, acceleration), determine the future motion. Therefore, how to effectively exploit  spatial and temporal cues to construct context-aware representations should be studied in depth to achieve physically reliable predictions.
\end{itemize}

To address these issues, we present an efficient model to learn physical dynamics from visual images and make object-centric predictions without any supervisory signals. 
Specifically, a perceptual module with autoencoder architecture is trained via reconstruction to learn object-centric representations in visual scenes and reconstruct images. The perceptual encoder decomposes images into disentangled spatial features and object representations with rich physical properties, such as position, orientation, etc.  In contrast to implicit features, such representations with explicit physical meaning aid in the incorporation of physical laws. Next,  to effectively leverage contextual information in both spatial and temporal dimensions, a context-aware aggregator is developed in the dynamic module, which constructs hybrid representations in vector form. Thus a good trade-off between object representation expressiveness and computational efficiency is achieved. Moreover, an interaction-aware predictor learns interactions between objects to predict the future state trajectory. Finally, the perceptual decoder combines the predicted states and spatial features to synthesize future images. 
% The proposed approach is divided into three main steps. First, perceptual encoder decomposes the images into spatial features and state representations of each object. Then, dynamic module aggregates environmental and temporal information to construct context-aware hybrid representations and predict  object states at the next moment. Finally, perceptual decoder combines the predicted states and spatial features to synthesize future frames in pixel space. 
Our main contributions are summarized as follows:
\begin{itemize}
\item [(1)] We present a general framework of object-centric prediction methods for visual dynamics learning, which focuses on object-centric representation and enables future predictions in state space from a combinatorial perspective.
\item [(2)] Based on this framework, we propose an unsupervised, context-aware model that extracts physically meaningful object representations from images and combines spatial and temporal contextual information to enhance the predictive ability.
\item [(3)] We conduct extensive experiments on several physical datasets to demonstrate the proposed model's effectiveness. The experimental results show that our approach achieves competitive performance with state-of-the-art methods and can generate more physically plausible predictions with good generalization ability.
\end{itemize}

This paper is organized as follows. Section II introduces the related work. Section III describes the overall network architecture and elaborates on the key modules. Section IV shows quantitative and qualitative experimental results compared to state-of-the-art methods on multiple physical scenarios. In the final section, we will draw a conclusion.

% \begin{figure}[ht]  %允许各个位置
%  \center{        
%  \includegraphics[scale=0.34]{new_pics/00_schematic_map.png}}  %缩放比例的倍数
%  \caption{Schematic diagram of non-object-centric and object-centric prediction methods. The former focuses on pixel-wise dense prediction and directly predicts the RGB value of pixels for future images. The latter is more concerned with the dynamics of objects in the visual scene. 
% }
%  \label{Schematic diagram}
% \end{figure}
\section{Related Work}
In this section, we give a brief review of the existing visual prediction methods and divide them into two categories: non-object-centric and object-centric methods, in terms of whether they rely on object-centric representation. Additionally, we also present context-aware prediction methods that take into account contextual information.
%, as depicted in Fig. \ref{Schematic diagram}

\begin{table*}[htbp]
    \renewcommand{\arraystretch}{1.3}
    \caption{Comparison of some representative object-centric prediction works, which predict the future from raw image inputs.  These models are described in terms of (1) the form of object-centric state representation, (2) whether object-level annotations are required and the type of supervisory signal, (3) whether contextual features, such as environmental information, are considered in the prediction process, and (4) whether having the ability to convert predicted states to visual images.}
    \centering
    \label{Comparison of some representative object-centric}
    \setlength{\tabcolsep}{4.5mm}{\begin{tabular}{c c c c c }
    \hline
    \hline
       Model &  Object-centric representation & Annotation-free    & Context-aware & Image generation \\ 
    \hline
    \multicolumn{5}{c}{Supervised methods}  \\ 
    \hline
        
    CVP \cite{ye2019compositional}         & 1D vector     & × (Positions)                     & ×                 & \checkmark                 \\ 
    RPIN \cite{qi2020learning}     & 3D feature map         & × (Bounding boxes)           & \checkmark    & ×                               \\ 
    PLATO \cite{piloto2022intuitive}            & 1D vector    & × (Masks)                           & ×                 &   \checkmark             \\ \hline
\multicolumn{5}{c}{Unsupervised methods}                                                                \\ \hline
 
Struct-VRNN \cite{minderer2019unsupervised}       &   1D vector & \checkmark     & ×   & \checkmark          \\ 
Grid keypoint \cite{gao2021accurate}   &   2D binary map & \checkmark           & ×   & \checkmark          \\ 
OPA \cite{schmeckpeper2021object}   &   Bounding boxes/masks/feature maps & \checkmark     
     & ×   & \checkmark          \\ 
Ours        &   1D vector   & \checkmark   & \checkmark   & \checkmark          \\ 
    \hline
    \hline

    \end{tabular}}
    
\end{table*}

\subsection{Non-object-centric prediction methods}

Non-object-centric prediction methods directly predict pixels in high-dimensional
image space. They usually extract high-level representations
from pixel inputs by convolutional units and model the dynamics by recurrent units \cite{wang2022predrnn}. Many efforts focus on various elaborate convolutional recurrent units to improve the ability to capture spatiotemporal evolution \cite{shi2015convolutional}, \cite{wang2017predrnn}, \cite{wang2019eidetic}. These models greatly increase the computational complexity while improving performance. To alleviate this problem, some works have designed more flexible and efficient structures by decomposing prediction space into motion and context \cite{villegas2017decomposing}, \cite{yu2019efficient}, \cite{wu2021motionrnn}. Some attempts incorporate physical knowledge in latent space to enhance prediction models \cite{guen2020disentangling}, \cite{zikri2023phylonet}. In addition to improvements in network architecture, delicate loss functions \cite{johnson2016perceptual} or complicated training strategies \cite{bengio2015scheduled}, \cite{chen2020long} are utilized to generate sharper predictions.  Moreover, deep stochastic prediction methods consider the inherent randomness of videos by incorporating uncertainty through latent variables \cite{babaeizadeh2018stochastic}, \cite{chen2020long}. Although these approaches are dedicated to modeling and predicting spatiotemporal variation in pixel space, they have been plagued by the difficulty of providing physically reliable predictions. Instead, our model follows an object-centric paradigm and learns physical dynamics behind visual scenes in state space.

\subsection{Object-centric prediction methods}
Object-centric visual prediction starts from the combinatorial nature of visual scenes and converts the pixel-to-pixel prediction task into a low-dimensional object-level state-to-state prediction. It mainly consists of three stages: (1) decompose visual scene into multiple objects and extracting physically meaningful representations of each object, (2) learn multi-object dynamics and inferring each object's future representation, (3) (sometimes) map predicted per-object representation to pixel space and synthesize future frames in a compositional way. In terms of whether labels are required, they fall into the following two categories. Table \ref{Comparison of some representative object-centric} summarizes the characteristics of representative object-centric  prediction models.

\subsubsection{Supervised methods}
Wu \textit{et al.} \cite{wu2017learning} train a recognition network to obtain the properties of each object from images, to recover physical states of visual scenes. A physics and graphics engine are utilized to simulate future states and render images.
Watters \textit{et al.} \cite{watters2017visual} propose visual interaction network, which obtains object-centric state codes from three consecutive frames and predicts the position/velocity vector. Ye \textit{et al.} \cite{ye2019compositional} use annotated 2D position and  deep feature extracted by DNNs as per-object representation. A decoder is designed to generate images from multiple latent object representations. O2P2 \cite{janner2019reasoning} requires segmentation masks of objects for object representations and a render module maps predicted representations to pixels. Baradel \textit{et al.} \cite{baradel2019cophy} take ground truth 3D object position and train a model to detect objects from visual images. Wu \textit{et al.} \cite{wu2020future} decompose the background scene and the foreground motion to synthesize future video. A segmentation network is trained by the annotation of moving objects to generate binary masks. Qi \textit{et al.} \cite{qi2020learning} leverage a region proposal network to build object representations. Piloto \textit{et al.} \cite{piloto2022intuitive} compress ground truth segmentation mask of each object to corresponding object codes by ComponentVAE. Although these supervised methods are effective for visual prediction tasks, they rely on massive labels, whose high cost limits their practical applications. In contrast, our model is an unsupervised approach that obtains object representations without annotation while achieving prediction performance comparable to supervised methods. 

\subsubsection{Unsupervised methods}
To avoid time-consuming video annotations, A few efforts are made to implement object-centric video prediction in an unsupervised manner. 
Byravan \textit{et al.} \cite{byravan2018se3}  uses the depth information provided in 3D point cloud to predict object masks and reason about object motion after manipulation for robotics control.
Zhu \textit{et al.} \cite{zhu2018object} estimate object masks from images and predict the dynamics of objects. 
Recently, KeypointNet\cite{jakab2018unsupervised} is widely used to extract per-object representation as a keypoint without supervision. Minderer \textit{et al.} propose Struct-VRNN to learn dynamics for stochastic video prediction and allow predictions conditioned on the input of agent's actions. Li \textit{et al.} \cite{li2020causal} leverage Transporter \cite{kulkarni2019unsupervised} to get keypoint representations of images and discover the causal relationship between objects from video. Gao \textit{et al.} \cite{gao2021accurate} represent keypoints as 2D binary maps instead of continuous spatial coordinates and convert a regression prediction task to a classification task avoiding accumulated errors. Janny \textit{et al.} \cite{janny2021filtered} learn hybrid latent keypoint-based representation and make counterfactual predictions when the initial situations change. Besides, Schmeckpeper \textit{et al.} \cite{schmeckpeper2021object} generate the pseudo-ground-truth mask of each object by an optical flow method, and an instance segmentation model is trained to obtain bounding box and mask as object representation. While most unsupervised methods do not consider the context of objects, our model exploits contextual information in environmental features and past trajectories, and fuses jointly in both spatial and temporal dimensions to obtain a context-aware hybrid representation, thus effectively improving the predictive ability.

 \begin{figure*}[htbp]  %允许各个位置
 \center{        
 \includegraphics[scale=0.48]{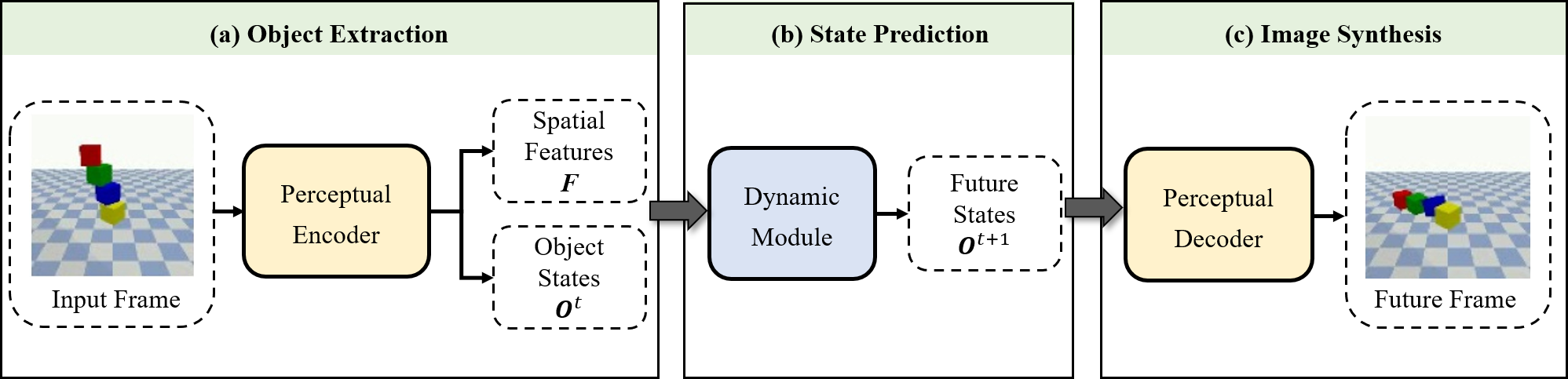}}  %缩放比例的倍数
 \caption{The overall architecture of the proposed model for unsupervised object-centric visual prediction. The perceptual module and dynamic module are shown in Fig. \ref{perceptual_module} and Fig. \ref{dynamic_module}. Detailed structures of them are described in Sec. \ref{Perceptual Module} and \ref{Dynamic Module}. The prediction procedure of our model consists of three stages: (a) The visual image is decomposed into spatial features $\textbf{F}$ and multiple physically meaningful object states $\textbf{O}^{t} = {\textbf{o}^t_{1:N}}$ in an unsupervised manner. (b) The dynamic module learns the underlying dynamics from past state trajectories and makes object-wise future predictions in state space. (c) Future frame in pixel space is produced by combining spatial features and predicted states. 
 }
 \label{Overall_Architecture}
\end{figure*}

\subsection{Context-aware prediction methods}
Contextual information has been shown in the literature to be beneficial for many vision tasks \cite{kampffmeyer2018connnet}, \cite{chen2023cpp}. Particularly, for prediction tasks, many context-aware methods have been proposed to achieve accurate predictions by taking spatiotemporal context into account.
% Contextual information, including interaction with the environment and objects, plays a crucial role in prediction tasks. Many context-aware methods have been proposed. 
Corona \textit{et al.} \cite{corona2020context-aware} constructs a directed semantic graph with objects and people in the scene as nodes, and learns interactions between nodes through graph attention networks. LaPred \cite{kim2021lapred} extracts contextual features from  trajectories of surrounding agents and lanes to predict the future motion of target agent. TNT \cite{zhao2021tnt} obtains elements such as agent trajectories and roads from HD maps and uses graph neural networks to encode the environmental context. AgentFormer \cite{yuan2021agentformer} presents an agent-aware attention mechanism to model dependency between multiple agents while obtaining the agent's corresponding contextual features from an annotated semantic graph. To better preserve  spatial information, some works use feature maps to represent the context information. RPIN \cite{qi2020learning} extracts the whole image features using a convolutional neural network and obtains feature maps containing rich contextual information based on bounding boxes via ROIPooling. Muse-VAE \cite{lee2022muse-vae} aligns local semantic map and trajectory represented by Gaussian heatmap to learn a joint representation of environment and agent motion for long-time trajectory prediction. Inspired by these works, we design a context-aware model that does not require additional information, like semantic maps. When the trajectory of an object is represented by Gaussian maps, our model focuses on local information about regions near objects. Graph neural networks are used to model interactions between objects and others or the environment. (see Sec. \ref{Dynamic Module}).
 
\section{Methodology}
In this section, we will present the overall architecture of the proposed model, followed by the detailed structure of its key modules.

\subsection{Overall architecture }
The overall architecture, as shown in Fig. \ref{Overall_Architecture}, is comprised of two modules, a perceptual module and a dynamic module. The prediction procedure for our object-centric visual prediction consists of three stages. First, the perceptual module encoder generates object-centric representations from images as the input for the dynamic module, i.e., the dynamic states of each object in visual scenes. Following that, the dynamic module constructs context-aware object representations, models physical dynamics from the previous trajectories of object states and makes predictions of per-object state at future timesteps. Finally, the perceptual module decoder combines predicted states and spatial features to produce future frames in pixel space.

\subsection{Perceptual Module}\label{Perceptual Module}
The perceptual module aims to decompose the continuous visual data into structured state representations, which include several attributes of the entity, such as location, appearance, and so on. KeypointNet and its variants \cite{kulkarni2019unsupervised}, \cite{jakab2018unsupervised}, \cite{dundar2021unsupervised} can extract keypoint-based representations from raw visual images in an end-to-end manner without any supervised signals. Based on the improved KeypointNet proposed in \cite{janny2021filtered}, we design a perceptual module for object extraction and image synthesis. In comparison to vanilla KeypointNet, it introduces additional orientation information to enhance the expression ability of object representation, thus improving the quality of generated images and facilitating the learning of dynamics.

\begin{figure}[htbp]  %允许各个位置
 \center{         
 \includegraphics[scale=0.43]{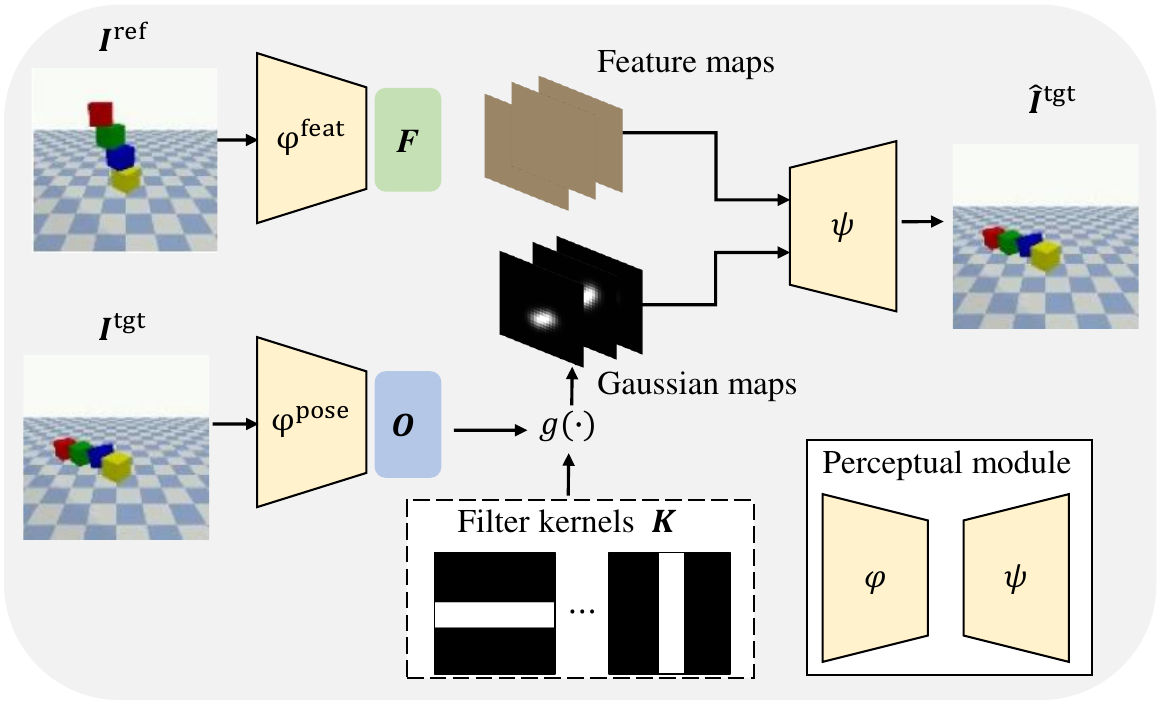}}  %缩放比例的倍数
 \caption{The schematic diagram of perceptual module. The encoder $\phi$ is used to obtain keypoints and static feature maps, while the decoder $\psi$ is used to reconstruct images. $\bm{F}$ and $\bm{O}$  denote spatial features and pose vectors of individual object.}
 \label{perceptual_module}
 \end{figure}

As shown in Fig. \ref{perceptual_module}, the perceptual module is an autoencoder architecture and consists of two CNN-based components, the encoder $\varphi$ and the decoder $\psi$. Given a video sequence $\bm{I}^{1:T} $, we randomly sample two frames at different timesteps as reference frame $\bm{I}^{\text{ref}}$  and target frame $\bm{I}^{\text{tgt}}$.  The target frame is reconstructed by the perceptual module through the following three steps.

\subsubsection{Object extraction} Input an image $\bm{I}$ with the size $H \times W$, and  assume $N$ keypoints in the image,  each keypoint's pose vector $\bm{o}=(\bm{p},\bm{s},\bm{r})$ includes 2d position $\bm{p} = (x,y)$, 1 scale coefficient $\bm{s}$, and $C$ rotation coefficients $\bm{r}$. The three-branch encoder network is utilized to extract disentangled $N$ pose vectors $\bm{o}_{1:N}= \varphi ^{\text{pose}} (\bm{I})  \in \mathbb{R}^{N \times {(3+C)}} $ and  appearance feature maps  $\bm{F} \in \mathbb{R}^{C^{\prime} \times H^{\prime} \times W^{\prime}}$, the former describing dynamic states of each keypoints and the latter containing static environment and appearance information. 
\begin{equation}
    \begin{aligned}
    \label{decompose_images}
\bm{F}=\varphi^{\text{feat}}(\bm{I}),\\
    \bm{p}_{1: N}=\varphi_{\text{pos}}^{\text{pose}}(\bm{I}),\\
    \bm{s}_{1: N},\bm{r}_{1: N}=\varphi_{\text{coef}}^{\text{pose}}(\bm{I}).
\end{aligned}
\end{equation}
In particular, keypoint position detector $\varphi_{\text{pos}}^{\text{pose}}$ outputs $N$ heatmaps $\bm{H}_{1:N} $,  then normalizes them via spatial softmax, as calculated by the following equation :
\begin{equation}
    \begin{aligned}
    \overline{\boldsymbol{H}}_n(i, j)=\frac{\exp \left(\boldsymbol{H}_n(i, j)\right)}{\sum_{i=1}^{H^{\prime}} \sum_{j=1}^{W^{\prime}} \exp \left(\boldsymbol{H}_n(i, j)\right)},
\end{aligned}
\end{equation}  
A weighted summation along the coordinates is then performed, thus compressing heatmaps into 2D coordinates:
\begin{equation}
    \begin{aligned}
    \bm{p}_n =\left(x_n, y_n\right)=\sum_{i=1}^{H^{\prime}} \sum_{j=1}^{W^{\prime}} \boldsymbol{u}(i, j) \cdot \overline{\boldsymbol{H}}_n(i, j),    
\end{aligned}
\end{equation} 
where $\bm{u}(i,j) \in \mathbb{R}^2 $ is the center coordinate of grid $(i,j)$.

\subsubsection{Gaussian-like maps construction} The process $g$ converts 1D pose vector into corresponding 2D Gaussian-like maps $ \bm{G} = g\left(\bm{o}_{1: N}\right) \in \mathbb{R}^{(N \times C) \times H^{\prime} \times W^{\prime}}$. First, we generate an isotropic Gaussian-like map that follows a Gaussian distribution centered at 2d coordinates $\bm{p}$ in space. Then the intensity of this map is scaled by scale factor $\bm{s}$ :
    \begin{equation}
    \begin{aligned}
    \overline{\bm{G}}_n=\bm{s}_n \cdot \exp \left(-\frac{1}{2 \sigma^2}\left\|\bm{u}-\bm{p}_n\right\|^2\right),
    \label{generate_gaussian}
    \end{aligned}
    \end{equation}
where $\sigma$ is a fixed value as the standard deviation of Gaussian distribution. Generally speaking, the scale factor $\bm{s}$ determines the size of objects in pixel space. Finally, the isotropic Gaussian map $ \overline{\bm{G}}_n$ is deformed by various filter kernels of different rotation angles and then weighted by the rotation coefficients $\bm{r}$ :
\begin{equation}
    \begin{aligned}
    \bm{G}_n=\left[\bm{G}_n^1, \ldots, \bm{G}_n^C\right], \bm{G}_n^i=\bm{r}_n^i \cdot\left(\bm{K}^i * \overline{\bm{G}}_n\right),
    \end{aligned}
    \end{equation}
where  $\bm{K}$ is the bank of filter kernels, as shown in Fig. \ref{generate_gaussian}. $\bm{K}^i$ is the filter kernel corresponding to the $i^{th}$ rotate coefficient. $*$ and $[ \cdot ]$ denote convolution and concatenation operation. 
\begin{figure}[htbp]  %允许各个位置
 \center{        
 \includegraphics[scale=0.53]{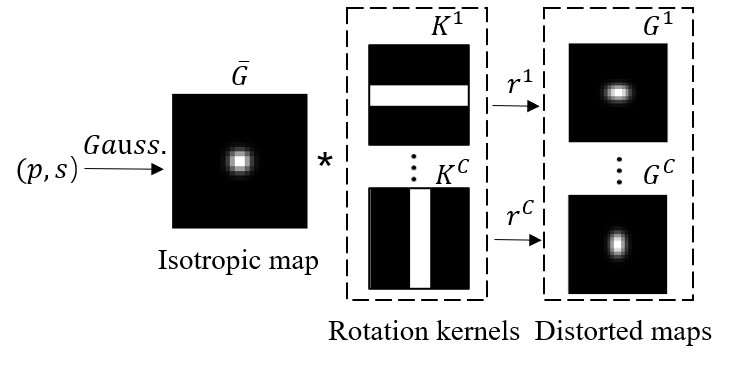}}  %缩放比例的倍数
 \caption{The illustration of Gaussian-like maps construction process $g$.}
 \label{gussain_construction}
 \end{figure}
 
\subsubsection{Image synthesis} Decoder network concatenates the appearance feature map from reference frame $\bm{F}^{\text{ref}}=\varphi^{\text{feat}}(\bm{I}^{\text{ref}})$ and Gaussian-like maps from target frame $\bm{G}^{\text{tgt}}=g(\bm{o}_{1:N}^{\text{tgt}})=g(\varphi^{\text{pos}}(\bm{I}^{tgt}))$  to reconstruct target frame:
\begin{equation}
\label{synthize_images}
\hat{\bm{I}}^{\text{tgt}}=\psi\left(\left[\bm{F}^{\text {ref}}, \bm{G}^{\text{tgt}}\right]\right).
\end{equation}
The decoder is implemented by stacking multiple convolutional and bilinear upsampling layers.

The encoder and decoder are jointly trained by reconstruction error. In addition to pixel-wise $L_2$ loss, gradient difference loss is introduced as a regularization term, to improve the sharpness of reconstruction:
\begin{equation}
L_{per}=\left\|\boldsymbol{\bm{I}}^{\text{tgt}}-\hat{\boldsymbol{\bm{I}}}^{\text{tgt}}\right\|_2^2+\lambda\left\|\nabla \boldsymbol{\bm{I}}^{\text{tgt}}-\nabla \widehat{\boldsymbol{\bm{I}}}^{\text{tgt}}\right\|_2^2,
\end{equation}
where $\boldsymbol{I}^{\text{tgt}}$ is target frame and $\widehat{\bm{I}}^{\text{tgt}}$ is the reconstruction. $\lambda$ is a hyperparameter and $\nabla (\cdot)$ denotes the sobel operator. 

\begin{figure*}[htbp]  %允许各个位置
 \center{        
 \includegraphics[scale=0.6]{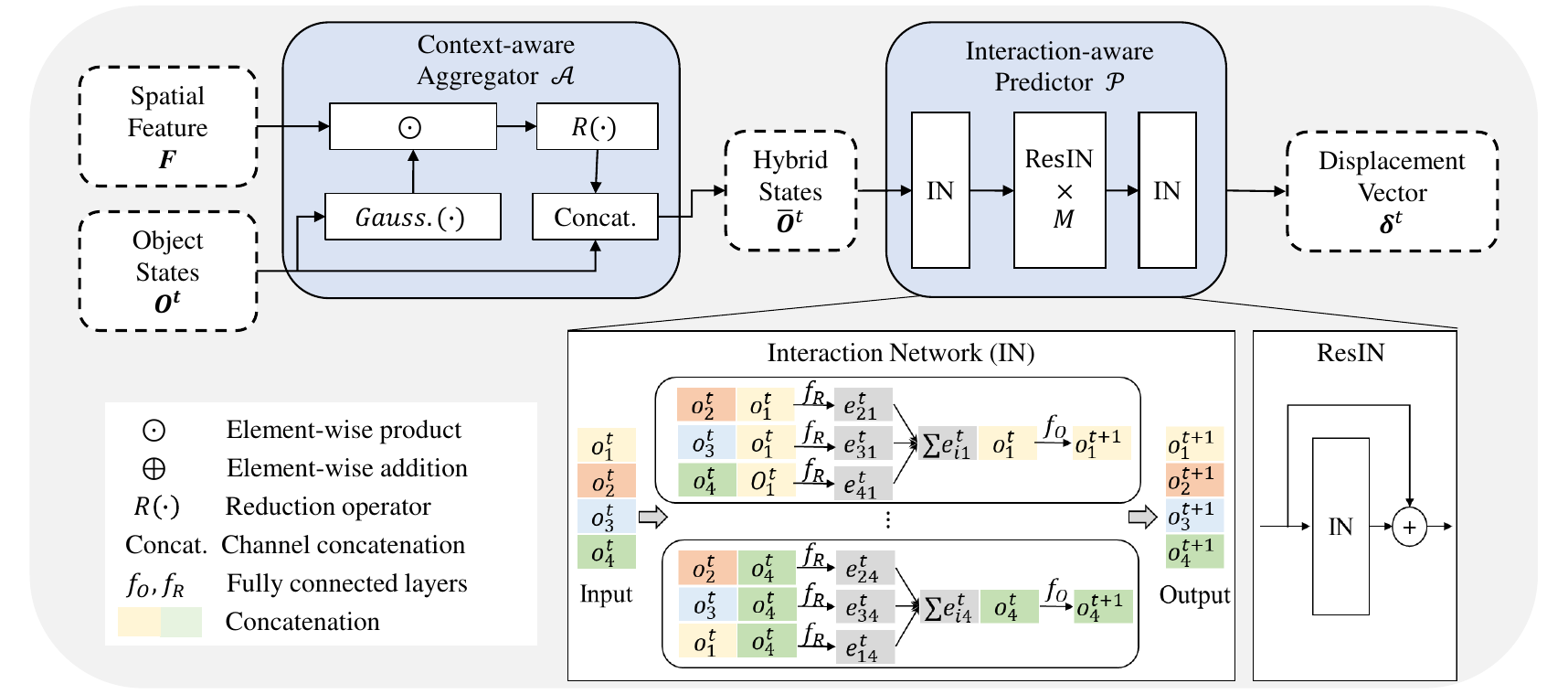}}  %缩放比例的倍数
 \caption{The schematic diagram of dynamic module  $ \mathcal{D}$. Context-aware aggregator $ \mathcal{A}$ aggregates static appearance information in feature map $F$ and the object's state $\bm{O}$ to a generate new state vector $\overline{\bm{O}} $, which additionally contains the contextual features around objects. Interaction-aware predictor $\mathcal{P}$ outputs the displacement vector to predict future states. Both of them are described in Sec. \ref{Context-aware aggregator} and Sec. \ref{Interaction-aware Dynamic predictor}.}
 \label{dynamic_module}
 \end{figure*}
The trained perceptual module is capable of keypoint-based pose extraction and image reconstruction, allowing us to generate temporally consistent keypoints from videos and transforming high-dimensional image sequences $\bm{I}^{1:T}$  into low-dimensional object-centric state sequences $\left\{\bm{o}_{1: N}^t\right\}_{t=1}^T= \varphi^{\text{pose}}(\bm{I}^{1:T})$. In our proposed prediction framework, the number of keypoints is set to the number of objects in the videos, thus each keypoint is treated as an object of interest. Previous consecutive frames $\mathcal{I}=\{\bm{I}^t\}_{t=1}^T$ are sequentially inputted into the pre-trained encoder $\varphi$ to obtain a temporally consistent state sequence for each object $\{\bm{O}^t\}_{t=1}^T=\{\{\bm{o}_n^t\}_{n=1}^N\}_{t=1}^T=\{\varphi^{\text{pose}}\left(\bm{I}^t\right)\}_{t=1}^T$. Due to the tracking nature of keypoint extractor, we don’t need any post-processing like object indexing. After dynamic module products future states of objects conditioned on past states, the decoder $\varPsi$ combines feature maps of 
the initial frame $\bm{F}=\varphi^{\text{feat}}\left(\bm{I}^1\right)$ and predicted object states to synthesize future frame  $\hat{\bm{I}}^{T+1}=\psi(\bm{F}, \hat{\bm{O}}^{T+1})$.

 \subsection{Dynamic Module}\label{Dynamic Module}
Our dynamic module $\mathcal{D}$ learns to understand the underlying dynamics of physical systems in order to forecast object states in the future. The structure of this module is shown in Fig. \ref{dynamic_module}, mainly consisting of a context-aware aggregator $\mathcal{A}$ and an interaction-aware predictor $\mathcal{P}$. Like a physical simulation engine, it generates the state trajectory of each object by iteratively performing single-step prediction:
\begin{equation}
\bm{O}^{t+1}=\mathcal{D}\left(\bm{O}^t, \bm{F}\right)+\bm{O}^t.
\end{equation}
\subsubsection{Context-aware aggregator}\label{Context-aware aggregator}
As mentioned above, the perception module provides a disentangled representation of a video frame, including several object dynamic states and spatial feature maps. The states evolve over time, while feature maps are shared in all frames. Obviously, the object states extracted directly lack environmental information, which plays an essential role in dynamics learning. To alleviate this problem, our dynamic module mines the contextual information encoded in the feature maps. In addition to object-object interaction, object-environment interaction and historical dynamic information are also introduced into the process of state prediction to jointly learn the dynamic evolution of physical systems and thus further improve the predictive power of the proposed model.

We design a context-aware aggregator $\mathcal{A}$ to fuse object states and context features around them in both spatial and temporal dimensions.  Given object states  $\bm{O}^t= \{\bm{o}_n^t \}_{n=1}^N$  at timestep $t$ and feature maps $\bm{F}$, gaussian maps are generated based on Eq. \ref{generate_gaussian}, where the closer the area is to object, the larger the value. Then, gaussian maps are used as spatial attention weights to reweight feature maps $\bm{F}$ of each channel dimension:
\begin{equation}
\bm{M}_n^t= \overline{\bm{G}}_n^t \odot \bm{F},
\end{equation}
where $\odot$ denotes Hadamard Product and $\bm{M}_n^t \in \mathcal{R}^{C^{\prime} \times H^{\prime} \times W^{\prime}}$ is the reweighted features. $\bm{M}_n^t$  is passed to a global reduction operator (such as SUM, AVG, or MAX) to aggregate environmental features and  condense reweighted feature maps $\bm{M}_n^t$ into a 1D latent vector $\bm{a}_n^t$, which contains the environmental information around the object.  

Since the object state has explicit physical properties, such as position, some prior knowledge can guide the prediction.  
In order to efficiently utilize hidden historical clues in position changes of successive frames, we compute
previous $\tau$ velocities by finite difference:
\begin{equation}
\begin{aligned}
\dot{\bm{p}}^t=\bm{p}^t-\bm{p}^{t-1}, 
\end{aligned}
\end{equation}
where $\bm{p}^t$ and $\dot{\bm{p}}^t$ are the position and velocity at timestep $t$.  The number of velocities $\tau$ determines the temporal receptive field of the dynamic module because  new states only contain dynamic information of the current frame and previous $\tau$ frames. It will be a trade-off between the temporal receptive field $\tau + 1$ and the effective length of new state sequences $T - \tau$, when given previous $T$ frames.  

We concentrate  latent vector $\bm{a}_n^t$, velocity sequence  with the original state $\bm{o}_n^t$ along channel dimension to achieve object-wise information aggregation and get a context-aware object-centric representation:
\begin{equation}
\begin{aligned}
\overline{\bm{o}}^t_n &= [\bm{o}_n^t,\dot{\bm{p}}^t_n,\ldots,\dot{\bm{p}}^{t-\tau+1}_n,\bm{a}_n^t]. 
% \\
% \overline{\bm{O}}^t&=\mathcal{A}\left(\bm{O}^t, \bm{F}\right).
\end{aligned}
\end{equation}

\subsubsection{Interaction-aware predictor} \label{Interaction-aware Dynamic predictor}
The predictor $\mathbf{P}$ aims to forecast future states from the hybrid object representation fed by aggregator $\mathbf{A}$. Our interaction-aware predictor $\mathcal{P}$ uses IN as a building block to learn the underlying physical dynamics behind videos. Interaction networks can access the physical states of nodes, reason about interactions, and have a deeper understanding of physical concepts. We first give a brief introduction to IN, and then describe the workflow of the predictor $\mathcal{P}$ to output the displacement vector of states $\bm{\delta}^t$ in the next timestep.

Interaction network \cite{battaglia2016interaction} is proposed as a learnable physics engine, which can model nonlinear dynamics and support forward simulation.  It describes a dynamic system as a directed graph composed of nodes and edges $G = <O,R>$, where nodes and edges represent objects and their relations in the system:
\begin{equation}
\begin{aligned}
&O=\left\{o_n\right\}_{n=1, \ldots,N},
\\
&R=\left\{r_m\right\}_{m=1, \ldots,M}, 
\end{aligned}
\end{equation}
where $N$ and $M$ are the number of objects and relations.
From a combinatorial perspective, IN designs an object-centric function $f_O$ and a relation-centric function $f_R$ to infer interaction between objects and update their states.
$f_R$ computes the effect of a relation:
\begin{equation}
e_m^{t}=f_R\left(\text {sender}_m^t, \text {receiver}_m^t, a_m^r\right),
\end{equation}
where $\text{receiver}_m^t$, $\text{sender}_m^t$ and $ a_m^r$ denote the receiver state, sender state, and attribute of this relation $r_m$. $f_O$ aggregates all effects of relations that point to object $n$ and predict the future:
\begin{equation}
o_n^{t+1}=f_O(o_n^t, \sum_{m \in P_n} e_{m}^t),
\end{equation}
where $P_n$ denotes the set of directed edges, whose receiver is node $n$. $e_m^t$ corresponds the effect of relation $r_m$ at time $t$.

We design ResIN block by adding a skip connection \cite{he2016deep} between the input and output of Interaction network, as shown in Fig. \ref{dynamic_module}. Its ResNet-like structure not only mitigates the potential problem of gradient vanishing but also empowers IN block as an approximator to the time derivative of states \cite{long2018pde-net}. The output of ResIN block can be calculated by the following:
\begin{equation}
\bm{O}^{t+1}=\operatorname{ResIN}\left(\bm{O}^t\right)=\operatorname{IN}\left(\bm{O}^t\right)+\bm{O}^t.
\end{equation}

Each node in the graph corresponds to an object representation in the physical system, which contains information about location, orientation, and surrounding environment. To simplify the learning of complex dynamics, the input states are embedded into a high-dimensional latent space  to approximately linearize nonlinear dynamics \cite{lusch2018deep}. The latent embeddings are sent to multiple stacked ResINs to infer the evolution of the dynamics. The predicted output is projected into a low-dimensional state space:
\begin{equation}
\label{next_step_prediction}
\delta^t=\mathcal{P}\left(\overline{O}^t\right),
\end{equation}
where $\delta^{t+1}$ reflects the physically meaningful increment of object states, such as velocity or acceleration.

During training, the dynamic module learns dynamics from observed state sequences and forecasts future states. this module is optimized by prediction error of states, which is defined as follows:
\begin{equation}
L_{dyn}=\sum_{t=T+1}^{T+\Delta T} \sum_{n=1}^N\left\|\bm{o}_n^t-\hat{\bm{o}}_n^t\right\|_2^2,
\end{equation}
where $\hat{\bm{o}}_n^t$ denotes predicted state vector of $n$th object in visual scene at $t$th frame and $\bm{o}_n^t$ is the ground truth state extracted by encoder $\varphi$. $N$ is the total amount of objects in the video. $T$ and $\Delta T$ are the length of historical frames and predicted frames. Alg. \ref{Psudo precedure} outlines the pseudo prediction procedure of our method in the inference stage.

\floatname{algorithm}{Algorithm}
\renewcommand{\algorithmicrequire}{\textbf{Input:}}
\renewcommand{\algorithmicensure}{\textbf{Output:}}
\begin{algorithm}
    \caption{Pseudo procedure of proposed method }
    \begin{algorithmic} %每行显示行号
        \Require Learned perceptual encoder $ \varphi(\cdot)$ and decoder $ \psi(\cdot)$ 
        \Statex \qquad $\text{Learned dynamic module} \ \mathcal{D}(\cdot)$
        \Statex \qquad $\text{Observed frames} \ \bm{I}_{\text{obs}} = (\bm{I}^1, ..., \bm{I}^T) \in \mathbb{R}^{T \times C \times H \times W }$
        \Statex \qquad $\text{Predicted horizon} \ \Delta T$
        \Statex \qquad $\text{Number of objects} \ N$
    
        \Ensure Future object representations  $\{\hat{O}^t\}_{t=T+1}^{T+\Delta T}$ 
        \Statex \qquad $ \text{Predicted frames} \ \hat{\bm{I}}_{\text{pred}} = (\hat{\bm{I}}^{T+1}, \cdots, \hat{\bm{I}}^{T+\Delta T})$  
        
        \State $ $
        \State $ \textbf{Phase 1: Decompose images into multiple objects}$
        \State $\text{Extract  feature maps and keypoints based on Eq. \ref{decompose_images}}  $
         \Statex \qquad \qquad $ \left\{\bm{F}, \bm{O}^t\right\}_{t=1}^T \leftarrow\left\{\varphi{(\bm{I}^t)}\right\}_{t=1}^T$
        % \State $ \textbf{Phase 1: Extract object-centric representation from images  }$
        
         \State $ \textbf{Phase 2: Make multi-step prediction}$
         \State $ \text{Initialize current states vectors} \ \bm{O}^{cur} \gets \bm{O}^1$
        \For{$t = 1 \to T+\Delta T-1$}
            \State $\text{Create new states by aggregator $\mathcal{A}$}$
        \Statex \qquad \qquad $ \{\overline{\bm{O}}^{cur}\} \gets \mathcal{A}(\bm{O}^{cur}, \bm{F})$
            \State $\text{Make next-step prediction} \  \hat{\bm{O}}^{t+1} \ \text{based on Eq. \ref{next_step_prediction}} $ 
            \Statex \qquad \qquad $ \hat{\bm{O}}^{t+1} \gets   \mathcal{P}(\overline{\bm{O}}^{cur}) + \bm{O}^{cur}$
            \State $\text{Update current states by predicted results} $
            \Statex \qquad \qquad $\ \bm{O}^{cur} \gets \hat{\bm{O}}^{t+1}$
           
        \EndFor
        
        \State $ \textbf{Phase 3: Produce future frames}$
        \State $\text{Synthesize images based on Eq. \ref{synthize_images} } $ 
        \Statex  \qquad \qquad $\{\hat{\bm{I}}^t\}_{t=T+1}^{T+\Delta T} \gets  \psi{(\bm{F}, \{\hat{\bm{O}}^t\}_{t=T+1}^{T+\Delta T})} $
        \State $ $
        \State \Return{$\{\hat{\bm{O}}^t\}_{t=T+1}^{T+\Delta T} \ \text{and} \  \hat{\bm{I}}_{\text{pred}} $}
    
    \end{algorithmic}
    \label{Psudo precedure}    
\end{algorithm}

In our work, we use the fully connected IN, which means that there are $n(n-1)$ relations for a system of $n$ objects. This is because objects may be in contact with arbitrary other objects as time unrolls. It is worth noting that we ignore stationary ground or walls because they can't usually be represented by a single keypoint. Instead, we introduce static environmental features into state prediction via a context-aware aggregator (see Sec. \ref{Context-aware aggregator}). 
% With the environment aggregator, static environment features are used to construct composite target representations that participate in subsequent state prediction.

\section{Experiments}

In this section, we first introduce the datasets, evaluation metrics, and comparison methods. We conduct experiments on multiple types of classical physical systems and compare our model with several state-of-the-art models. Ablation analyses demonstrate the effectiveness of key components of the proposed models.

\subsection{Experimental Setup}
\subsubsection{Datasets} We evaluate the predictive performance of models on several synthetic datasets of commonly used physical scenarios. The statistics of datasets and corresponding experimental settings are summarized in Table \ref{dataset_statistics}.

\begin{itemize}
\item \textbf{Blocktower and Balls scenario}: 
They are two scenarios of a simulated physics benchmark Filtered-Cophy \cite{janny2021filtered}. The above two scenarios are similar to physical systems commonly used for physical reasoning tasks, such as ShapeStacks \cite{groth2018shapestacks} or Billiards \cite{fragkiadaki2015learning}. Blocktower scenario features multiple stacked cubes, static in contact initially and potentially unstable under gravity. We predict $\Delta T=25$ frames conditioned on a single initial frame.  In Balls scenario, moving balls with different velocities collide with each other and bounce against the surrounding walls. Due to the initial velocity of each ball, the input of consecutive frames is necessary for future frame prediction. We set $T=3$ and $\Delta T=25$.  Experiments are conducted on the subsets where multiple objects have the same mass. This setting is mainly due to the fact that mass, as a confounder, is hardly inferred from a few observations.  
\item \textbf{Simb dataset}: This simulation billiard dataset is similar to Balls scenario, but has a simpler background. It has 1000 videos for training and 1000 videos for testing. Each video has 1000 frames. We perform only trajectory prediction task on this dataset. Following the experimental setup in \cite{qi2020learning}, we use a sliding window of stride 1 to split the video into many sub-clips and predict future object positions in the 20 future timesteps, conditioned previous 4 timesteps. 
\end{itemize}

\begin{figure}[htbp]  %允许各个位置
 \center{        
 \includegraphics[scale=0.43]{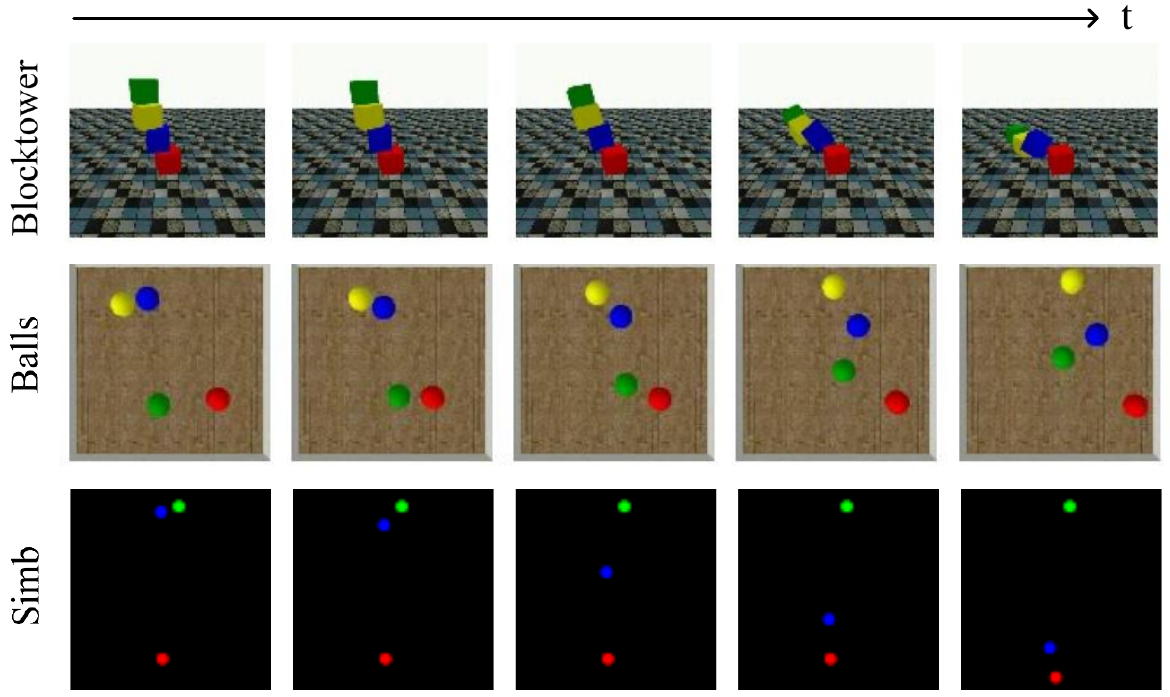}}  %缩放比例的倍数
 \caption{Visualization of datasets unrolling over time.}
 \label{vis_dataset}
 \end{figure}
 
\begin{table}[htbp]
\renewcommand{\arraystretch}{1.3}
    \caption{The statistics of datasets and corresponding experimental setting. $(C,H,W)$ is the size of each frame. During training, We predict $\Delta T$ future frames, given $T$ previous frames.}
    \centering
    \label{dataset_statistics}
    \setlength{\tabcolsep}{2.1mm}
\begin{tabular}{c c c c c c c }
\toprule
Scenario & $(C,H,W)$ & $N_{train}$ & $N_{val}$ & $N_{test}$ & $T$ & $\Delta T$ \\
\midrule
Blocktower & (3,112,112)&1486 & 384 & 188 & 1 & 25 \\
Balls & (3,112,112) &1734 & 375& 188& 3 & 25\\
Simb & (3,64,64) &1000 & 1000 & / & 4 & 20\\
 \bottomrule
\end{tabular}
\end{table}

\subsubsection{Evaluation Metrics} Although the prediction quality of object-centric methods in pixel space is often limited by their ability to synthesize images, for a fair comparison, we still chose several widely used video quality evaluation metrics to assess model performance. Following previous work, we adopt four commonly used metrics to evaluate the prediction performance of models from different aspects.
\begin{itemize}
\item \textbf{Mean Square Error (MSE)}: MSE between ground truth and the reconstructed image is a simple and commonly used metric, but often prefers blurry predictions.
\item \textbf{Structural Similarity Index Measure (SSIM)}: Luminance, contrast, and structure are the three factors that SSIM \cite{wang2004image} uses to measure the similarity between two images. 
\item \textbf{Local Peak Signal to Noise Ratio (L-PSNR)}: The standard PSNR measures similarity through pixel-wise comparison and may fail to capture temporal evolution because static background occupies most of the pixels. Consistent with previous work \cite{janny2021filtered}, we introduce L-PSNR, which only considers local regions near objects of interest. 
\item \textbf{Learned Perceptual Image Patch Similarity (LPIPS)}: LPIPS \cite{zhang2018unreasonable} leverages in-depth features, which deep neural networks extract from two images, to calculate perceptual similarity. It has been shown that LPIPS is more consistent with human perception than pixel-based evaluation metrics.
\end{itemize}

\subsubsection{Implement Details} We implement the proposed method with the Pytorch framework and conduct experiments on a single NVIDIA GeForce RTX 3090 GPU. With an initial learning rate of 5e-4, perceptual module and dynamic module are trained independently with a mini-batch of 16 using Adam optimizer. We reduce the learning rate to half when the validation error stops decreasing for over 10 epochs. For perceptual module, the number of appearance coefficients $C$ is 4. For dynamic module, the length of previous velocity sequence $\tau$ is set to 0 and 2 in Blocktower and Balls scenario. $f_O$ and $f_R$ are implemented by a multiple-layer perception with two hidden layers. In dynamic predictor $\mathcal{P}$, we stack 2 ResIN blocks for multi-step message passing. The size of latent embeddings is 256-dim. 
% Furthermore, we use a curriculum learning strategy, i.e., scheduled sampling, to mitigate data distribution inconsistency during the training and inference phases.

\subsection{Comparison with State-of-the-Art}
Future frame prediction and trajectory prediction tasks are used to evaluate the predictive ability of models. The difference between them is whether future images need to be generated or not. When we compare the proposed method with unsupervised baselines by evaluating the quality of predicted frames, we only perform comparison experiments with supervised baselines on the latter task, because most supervised object-centric methods lack the ability to generate images.

\subsubsection{Future Frame Prediction}
We compare our method with seven unsupervised  prediction baselines, including four non-object-centric methods and three object-centric methods. 
As non-object-centric approaches, ConvLSTM \cite{shi2015convolutional}, PredRNN\cite{wang2017predrnn}, PhyDNet \cite{guen2020disentangling} and SimVP \cite{gao2022simvp} belong to RNN-RNN-RNN, CNN-RNN-CNN, and CNN-CNN-CNN categories, which are representative architectures of video prediction works \cite{gao2022simvp}. Among the object-centric approaches, Struct-VRNN \cite{minderer2019unsupervised} and Grid keypoint \cite{gao2021accurate} are designed for stochastic video prediction, while V-CDN \cite{li2020causal} is a deterministic approach like our method. We reimplement these methods based on their officially released codes to obtain experimental results. In our experiments, all compared models are trained to predict $\Delta T $ future frames by observing past $T$  frames, summarized in Table \ref{dataset_statistics}. Especially, SimVP can only predict the future as the same length as the input sequence, so we iterative its prediction procedure to produce longer predictions. For V-CDN, as in previous work \cite{janny2021filtered}, we modify it to make it suitable for video prediction. For stochastic video prediction methods (Struct-VRNN and Grid keypoint), we follow their best-of-many prototype \cite{babaeizadeh2018stochastic}, where we choose the best one from many generated samples. We also report the model parameters and per-frame FLOPs on Balls scenario to evaluate their computational efficiency in Table \ref{Model parameter comparison}. 

\begin{table}[htbp]
    \renewcommand{\arraystretch}{1.3}
    \caption{Model parameter and per-frame FLOPs comparison on Balls scenario.}
    \centering
    \label{Model parameter comparison}
    \setlength{\tabcolsep}{3.2mm}{\begin{tabular}{c c c c}
\toprule
\multicolumn{2}{c}{Model} & Params.(MB)           & FLOPS(G) \\
\midrule
\multirow{4}{*}{\begin{tabular}[c]{@{}c@{}}Non-\\ object-centric\end{tabular}}
& ConvLSTM & 4.30 & 3.37   \\
& PredRNN & 24.56 & 19.25    \\
& PhyDNet & 1.91 & 1.79     \\
& SimVP & 13.35 & 10.56 \\
\midrule
\multirow{4}{*}{Object-centric}                     
& Struct-VRNN & 2.24  & 3.06 \\
& V-CDN & 8.83  & 2.40    \\
& Grid keypoint & 3.44 & 4.88   \\
& Ours & 7.18  & 2.26 \\
\bottomrule  
\end{tabular}}
\end{table}

\begin{table*}[htbp]
\renewcommand{\arraystretch}{1.3}
    \caption{Quantitative results of different predictive learning models on Blocktower (B) and Balls (B) scenario, averaged over the prediction horizon. $\Delta T $ is the prediction horizon in training time. $\uparrow$ ($\downarrow$) indicates the higher (lower) is better. The optimal(or suboptimal) results are \textbf{bolded} (\underline{underlined}). }
    \centering
    \label{comparison results}
    \setlength{\tabcolsep}{1.3mm}
    
\begin{tabular}{c|c|cccc|cccc}
\toprule
\multirow{2}{*}{Scenario}   & \multirow{2}{*}{Model} & \multicolumn{4}{c|}{$t \in [0,\Delta T]$} & \multicolumn{4}{c}{$t \in [\Delta T,2\times \Delta T]$} \\ 
&  & MSE$\times 10^{-3}$(↓)  & L-PSNR(↑)& SSIM(↑) & LPIPS$\times 10^{-2}$(↓)  &  MSE$\times 10^{-3}$(↓)  & L-PSNR(↑)& SSIM(↑) & LPIPS$\times 10^{-2}$(↓) \\ 
%\hline
\midrule
                   
\multirow{8}{*}{Blocktower} & ConvLSTM   & 10.10  & 30.53  & 0.935  &  3.51                                     &  \underline{41.89}   & \underline{22.92}  & 0.863  & 16.46  \\                      %\cline{2-10} 
                    & PredRNN    & \underline{10.02}   & \underline{31.16}   & \textbf{0.954} & \underline{2.55}                & 42.84   & 22.28   & \underline{0.885} & 14.73 \\ %\cline{2-10} 
                    & PhyDNet    & 25.09  & 24.78  & 0.627 & 28.74                 & 64.76   & 21.44  & 0.584 & 39.19 \\ %\cline{2-10} 
                    & SimVP     & 10.44   & 30.15   & 0.932  & 3.22                & 44.84    & 22.24  & 0.856  & 15.22 \\ %\cline{2-10} 
                     & Struct-VRNN    & 12.93    & 27.94  & 0.940  & 3.10               & \textbf{28.88}    & \textbf{23.82} & \textbf{0.886}  & \textbf{9.88} \\ %\cline{2-10} 
                    & V-CDN     & 10.62    & 30.63  & 0.939  & 2.96                & 52.87    & 22.93  & 0.880  & 16.62 \\ %\cline{2-10} 
                    & Grid keypoint     & 13.35    & \textbf{32.70}  & \underline{0.949}  & 3.15               & 54.03   & 22.22  & 0.863  & 17.73 \\ %\cline{2-10} 
                    & Ours      &  \textbf{9.87}    & 31.01  & 0.943 &      \textbf{2.50}      &   45.95 &  22.87 & \underline{0.885} &  \underline{14.20} \\% \hline
                    \midrule
\multirow{8}{*}{Balls}  & ConvLSTM  & 9.04    & 29.05   & 0.920  & 4.77                                    & 33.70   & 22.09   & 0.851  & 18.60 \\                    %\cline{2-10} 
                    & PredRNN   & \underline{7.72}    & \underline{29.95}   & \underline{0.935}  & \underline{3.99}               & \underline{29.99}   & \underline{22.68}   & \underline{0.866}  & \underline{16.92} \\ %\cline{2-10} 
                    & PhyDNet   & 10.76   & 28.39   & 0.900  & 6.02                & 35.76   & 21.94   & 0.830  & 19.92 \\ %\cline{2-10} 
                    & SimVP     & 11.62   & 28.12   & 0.890  & 6.09                & 40.74   & 21.00   & 0.785  & 20.22 \\ %\cline{2-10} 
                    & Strut-VRNN     & 34.72   & 22.05   & 0.856   & 17.89 
                & 37.80  & 21.52  & 0.844  & 18.85  \\ %\cline{2-10} 
                    & V-CDN     & 22.00   & 25.39   & 0.898  & 9.27                & 50.23   & 20.33   & 0.845  & 26.89 \\
                    & Grid keypoint     & 25.74   & 23.72   & 0.880  & 14.59                & 46.19   & 20.45   & 0.845  & 26.89 \\
                    & Ours      & \textbf{5.95}   & \textbf{31.30}   & \textbf{0.947}  & \textbf{1.90}                 & \textbf{26.07}  & \textbf{23.86}  & \textbf{0.896}  & \textbf{9.60} \\ 
                   \bottomrule
                   
\end{tabular}
\end{table*}

Table \ref{comparison results} shows the quantitative comparison with all metrics averaged over the predicted frames. 
As an unsupervised object-centric approach, the visual quality of  reconstructed images by  the perceptual module determines the upper bound of the prediction performance of the proposed model. Although extra  pixel errors are introduced when predicted object states are decoded into images,  Our model still achieves competitive results on pixel-based metrics (MSE, SSIM, L-PSNR) compared to other models, which are optimized directly using pixel-wise loss. Moreover, our model shows consistent superiority in perceptual metrics LPIPS.  For  Blocktower  scenarios, predicting the future conditioned on a single frame is uncertain. Struct-VRNN, as a stochastic prediction method, models the uncertainty by a random latent variable  and thus benefits from  inherent advantages. Among deterministic methods, PredRNN has the closest performance to ours, which has a huge number of parameters (nearly three times as many as ours) and high computational complexity. 
Our model achieves an optimal compromise between predictive performance and computational efficiency. Besides, we compare the performance of all models of predicting future images in longer horizon $ t \in \left[ \Delta T, 2\times \Delta T \right] $, shown in Table \ref{comparison results}. Consistent with previous results, our model still outperforms other models. Especially, the proposed model has a more obvious advantage in Balls scenarios, where fast-moving balls and frequent collisions require a deeper understanding of the dynamics. The results demonstrate our model achieves state-of-the-art performance and benefits from modeling complex dynamics behind physical systems.

\begin{figure}[htbp]
\centering
\subfigure[Frame-wise MSE] %子图片标题
{\includegraphics[width=0.9\linewidth]{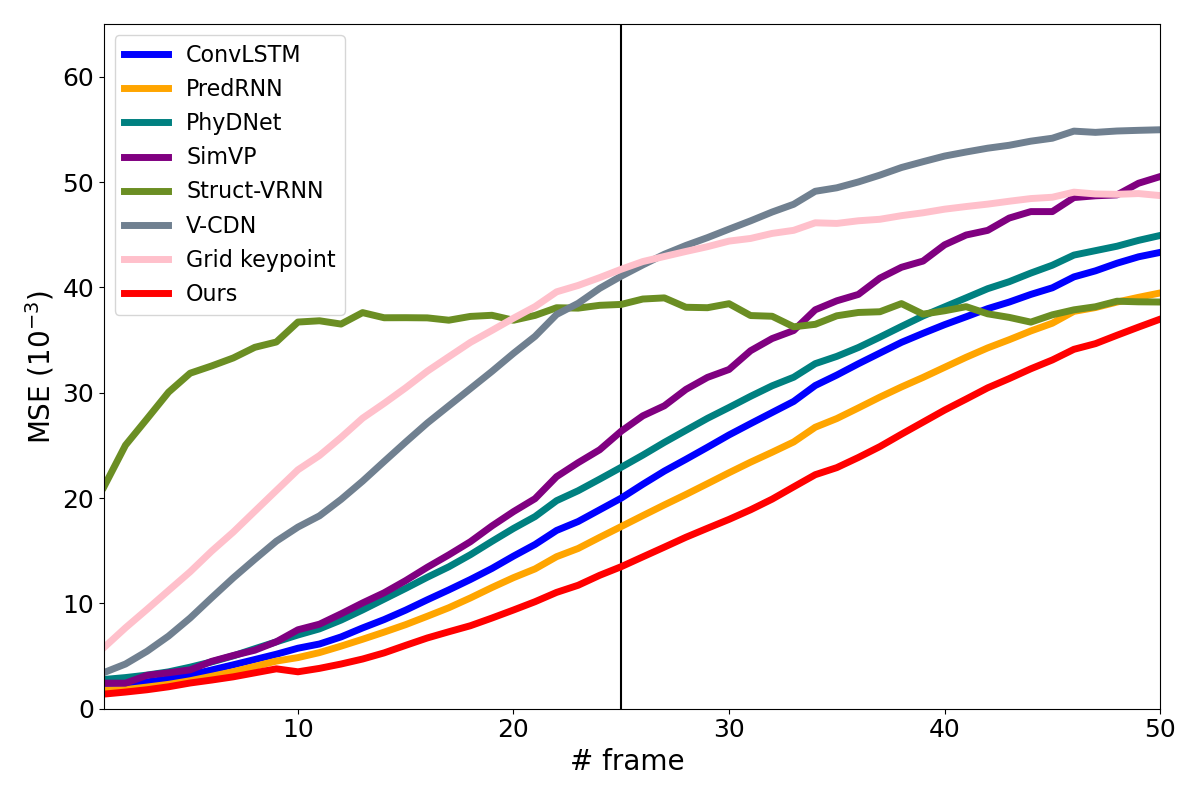}} %[图片大小]{图片路径}
\\
\subfigure[Frame-wise LPIPS]{\includegraphics[width=0.9\linewidth]{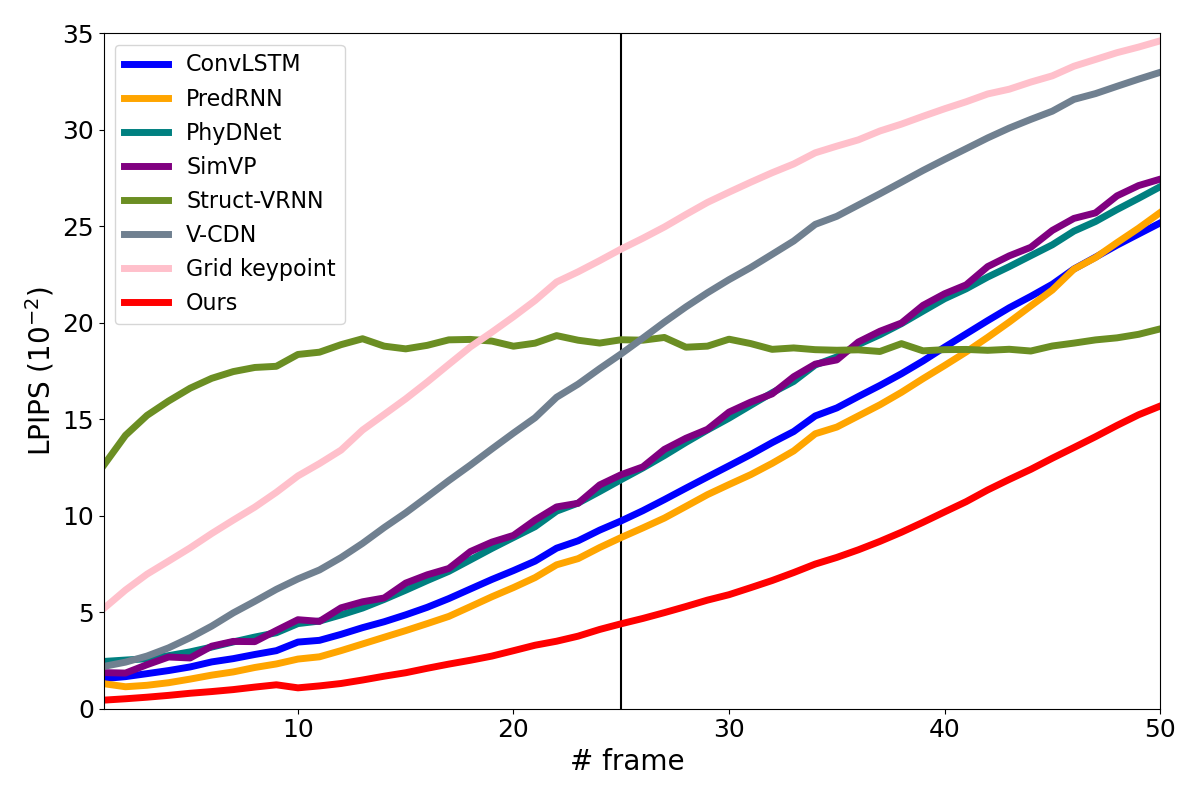}}
\caption{Frame-wise MSE and LPIPS comparisons of different models on Balls scenario. A lower MSE and LPIPS score denotes better performance. The solid line indicates the prediction horizon in the training phrase.} %图片标题
\label{LPIPS_per_frame}  %图片交叉引用时的标签
\end{figure}

% \begin{figure}[htbp]
% \centering
% \subfigure[Frame-wise MSE] %子图片标题
% {\includegraphics[width=0.49\linewidth]{new_pics/5_experiment_B_MSE.png}} %[图片大小]{图片路径}
% \subfigure[Frame-wise LPIPS]{\includegraphics[width=0.49\linewidth]{new_pics/6_experiment_B_LPIPS.png}}
% \caption{Frame-wise MSE and LPIPS comparisons of different models on Balls scenario. A lower MSE and LPIPS score denotes better performance. The solid line indicates the prediction horizon in the training phrase.} %图片标题
% \label{LPIPS_per_frame}  %图片交叉引用时的标签
% \end{figure}
For further analysis, we visualize frame-wise prediction quality of these models over predicted horizon in Fig. \ref{LPIPS_per_frame}.  
Our method reaches the state-of-the-art performance for both low-level pixel MSE metric  and high-level perceptual metrics LPIPS. For LPIPS metric, our model has maintained a large performance margin against other competitors in the full horizon. Also, we find that the deterioration of our method over time is slower than other deterministic methods and thus has a more significant advantage in long-horizon prediction. This experimental phenomenon validates our model's powerful predictive ability, particularly for interaction-intensive systems and longer horizons. 

\begin{figure*}[htbp]
\centering
\subfigure[Blocktower scenario] %子图片标题
{\includegraphics[width=0.49\linewidth]{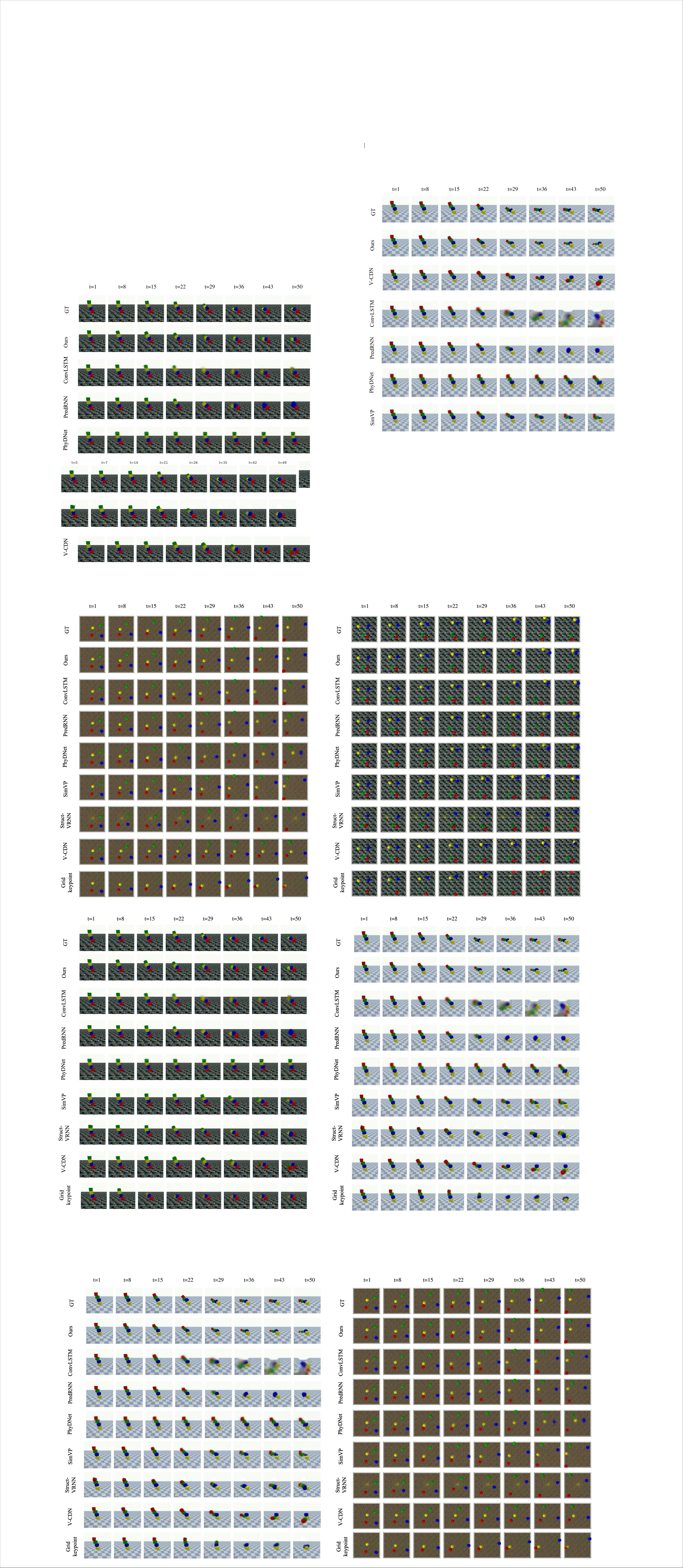}} %[图片大小]{图片路径}
\subfigure[Balls scenario]{\includegraphics[width=0.49\linewidth]{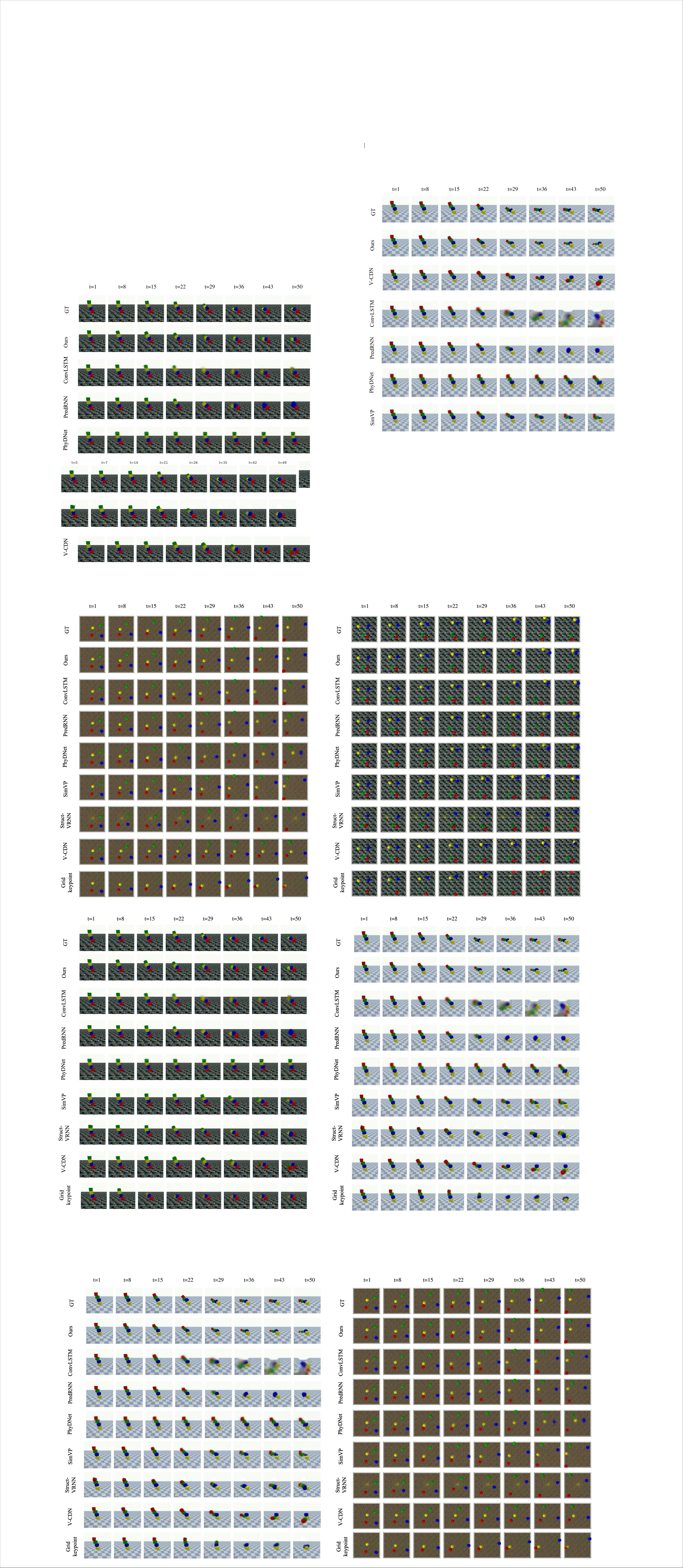}}
\caption{Visualization of predicted results on two scenarios. Compared to other competitors, our model makes more accurate predictions of objects' trajectories and retains more consistent details of object appearance in generated images} %图片标题
\label{visualization_prediction}  %图片交叉引用时的标签
\end{figure*}

 \begin{figure}[htbp]  %允许各个位置
 \center{        
 \includegraphics[scale=0.27]{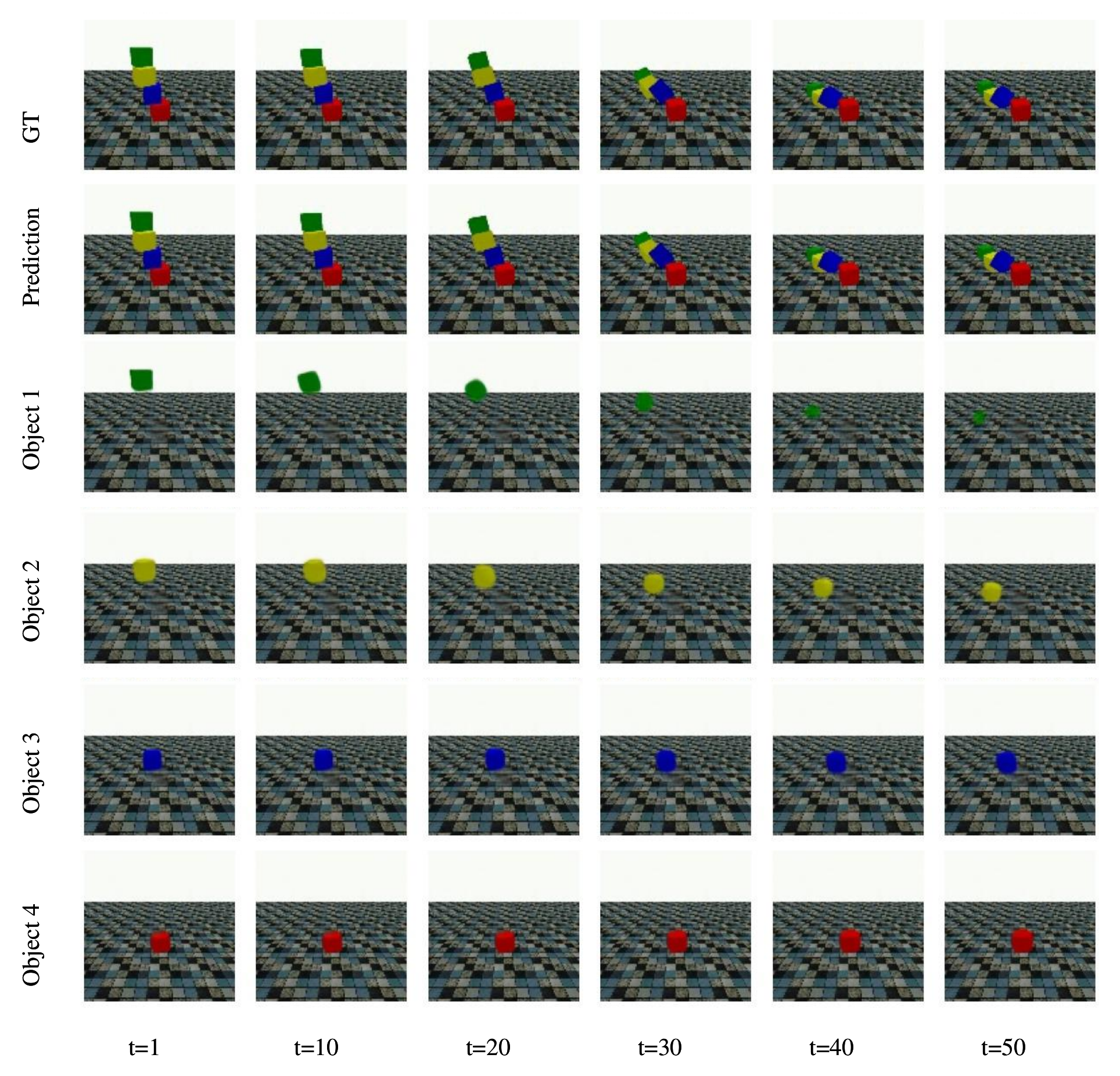}}  %缩放比例的倍数
 \caption{Visualization of individual object trajectory. Our model is able to make object-centric predictions and synthesize images based on predicted states of each object.}
 \label{object_path}
\end{figure}
Fig. \ref{visualization_prediction} shows qualitative comparisons with baselines. The visualization of prediction results demonstrates that our model has a deeper understanding of multi-object dynamics and captures more reliable temporal variations. Benefiting from  the context-aware part and interaction-aware part of the proposed dynamic module, our model makes a more reasonable trajectory when balls collide with walls or other balls.

A visualization example of predicted object trajectories is shown in Fig. \ref{object_path}. Our method generates the future trajectory of each object individually, thus allowing agents to manipulate objects for control and planning. We can also edit the properties (position or orientation) of objects to compose new scenes.  It's promising to make counterfactual predictions when agents intervene on a specific object. As the robot stacks cubes, trying to keep them stable, our model can predict the outcome of its actions and thus give reward signals to modify its decisions.

\subsubsection{Trajectory Prediction}
The goal of this task is to infer each object's position $(x,y)$ for $\Delta T$ future moments from $T$ past visual images, without involving future frame generation. Therefore, mean squared error between ground truth  and predicted positions is used as the evaluation metric.
We conduct experiments on the simulation billiard dataset to evaluate the proposed method's performance on the trajectory prediction task and compare it to several supervised object-centric methods, such as VIN \cite{watters2017visual}, CVP  \cite{ye2019compositional} and RPIN \cite{qi2020learning}. 
While these supervised algorithms rely on object-level annotation, the proposed method in this paper obtains the object state in an unsupervised manner as pseudo-labels and predicts trajectory based on this. The object's annotation is only used to measure the prediction error of position.

The quantitative results in Table \ref{simb results} show that the proposed unsupervised model achieves a powerful predictive power comparable to that of supervised methods. All of these models learn physical dynamics with IN as the building block and the main difference lies in the form of their object representation, vector or feature maps. As claimed above, we construct a hybrid object representation that includes not only position but also incorporates environmental features as well as historical velocities. This allows us to exploit a vector-form representation that retains more spatial and temporal information, contributing to more accurate trajectory prediction. The superiority of the proposed method for this trajectory prediction task further emphasizes the importance of a rich object representation, and the effectiveness of the perceptual module and context-aware aggregator proposed in Sec. \ref{Perceptual Module} and Sec. \ref{Context-aware aggregator}.

\begin{figure}[htbp]  %允许各个位置
 \center{        
 \includegraphics[scale=0.25]{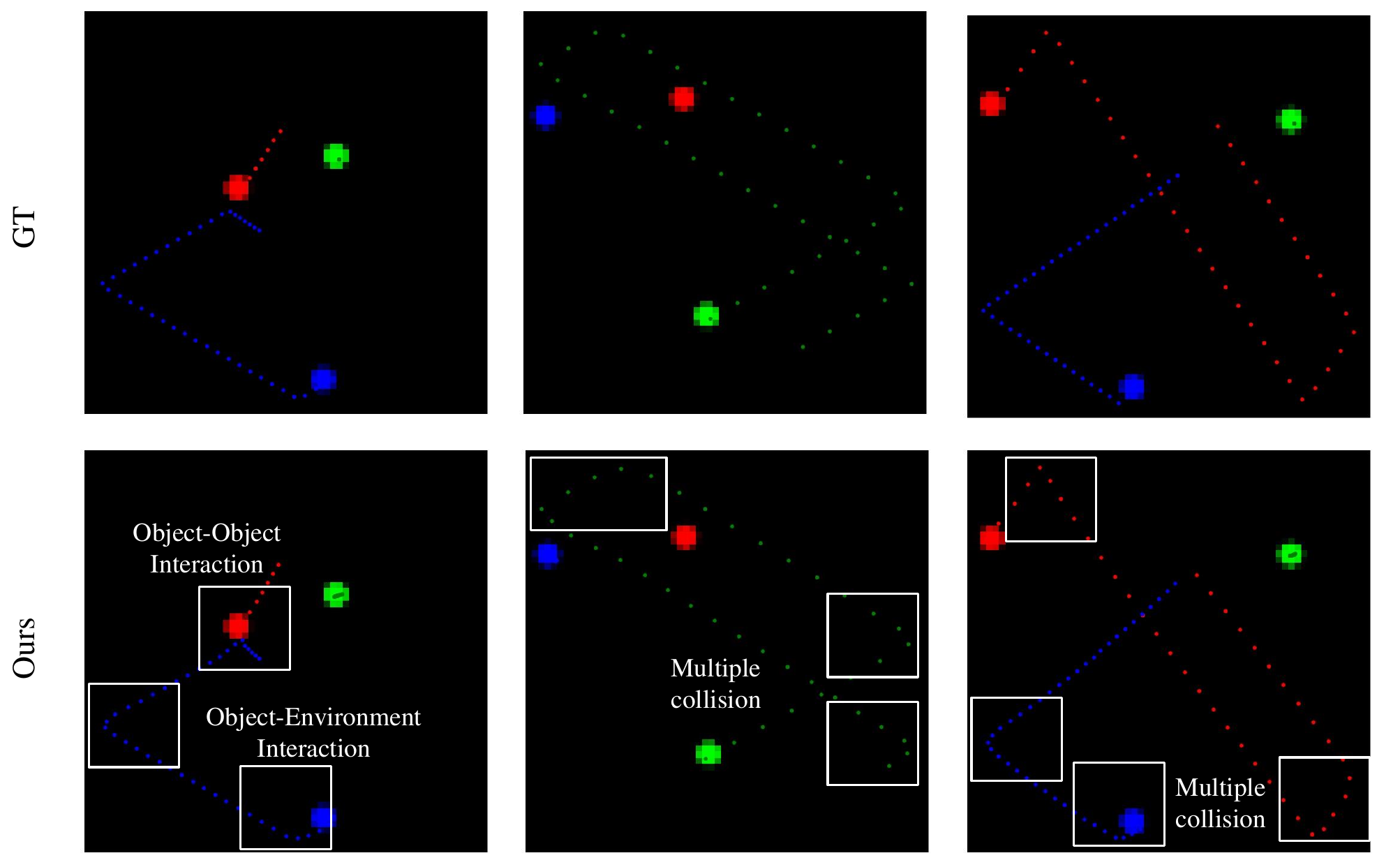}}  %缩放比例的倍数
 \caption{Visualization of ground truth and our predicted trajectory on Simb dataset.  Some of the collisions are depicted in the white boxes.}
 \label{vis_simb}
 \end{figure}

Qualitative results of our predicted trajectory are shown in Fig. \ref{vis_simb}. The proposed model can accurately infer object-object interaction and object-environment interaction, and generate physically plausible future trajectories (e.g., the bounce angle after collision with a wall, 
and movement of the stationary object after collision with moving others).
\begin{table}[htbp]
\renewcommand{\arraystretch}{1.3}
 \caption{Quantitative comparison on Simb dataset. We report the object representation form and  MSE $\times 10^{-3}$ (smaller is better) of the predicted trajectory. For supervised models, we used the results provided in \cite{qi2020learning}.}
    \centering
    \label{simb results}
    \setlength{\tabcolsep}{0.6mm}
\begin{tabular}{cc|c|c|c}
\toprule
\multicolumn{2}{c|}{Model}               & Object repre. & $t \in [0,\Delta T]$ &$t \in [\Delta T, 2 \times \Delta T]$   \\
\midrule
\multirow{4}{*}{Supervised} & VIN  & Vector   & 3.89  & 29.51   \\
                            & CVP  & Vector    & 80.01 & 108.56   \\
                            & RPIN (IN)  & Vector  & 3.01 & 27.88  \\
                            & RPIN (CIN) & Feature maps & 2.55 &25.77 \\
\midrule
Unsupervised                 & Ours     & Vector   &  \textbf{1.73} & \textbf{16.10} \\  
\bottomrule
\end{tabular}
\end{table}

\subsection{Ablation Study}

Both our proposed method and object-centric competitors adopt a keypoint-based model as the perceptual module. Thus their predictive capability is constrained  by  the reconstruction quality of keypoint representation. In Table \ref{Performance_degeneration_rate}, we report reconstruction and prediction performance, and the corresponding degradation rate. 
The results show that in addition to performance gain from our expressive perceptual module, the dynamic module in the proposed model achieves minimal performance degradation, which is superior to other competitors. Thanks to the structured architecture of object-centric methods, we can equip them with the same perceptual module and preserve their unique dynamic parts to compare more clearly the capabilities of different dynamic modules. In the following experiment, a perceptual baseline model is obtained by degrading our perceptual module to vanilla keypointNet, i.e., keypoints 
 are represented by position $(x,y)$ only. As we can see in Table \ref{results_with_base_perceptual_model}, shared the same object representation, our model still outperforms other object-centric methods. This phenomenon further demonstrates the effectiveness of the proposed dynamic module.
\begin{table}[htbp]
\renewcommand{\arraystretch}{1.3}
\caption{Performance degeneration rate of various dynamics module. We report  the reconstruction  prediction performance (SSIM) and  degeneration rate for $t \in [0,2 \times \Delta T]$ on Balls scenario. }
\centering
\label{Performance_degeneration_rate}
\begin{tabular}{cccc}
\toprule
Model                             & Recon. Perf.         & Pred. Perf. & Degeneration Rate \\
\midrule
Struct-VRNN                       & 0.931                    & 0.850           &    8.7$\%$               \\
V-CDN                             & 0.970                    & 0.871           &   10.2$\%$                \\
Grid keypoint &  0.937 &    0.858         &    8.5$\%$               \\
Ours     & 0.978        & 0.922          &   \textbf{5.7$\%$} \\   \bottomrule            
\end{tabular}
\end{table}

\begin{table}[htbp]
\renewcommand{\arraystretch}{1.3}
\caption{Quantitative comparisons of different models with the same perceptual baseline on Balls scenario. We report  prediction error of object position and quality of future frames for $t \in [0, \Delta T]$. }
\centering
\label{results_with_base_perceptual_model}
\setlength{\tabcolsep}{1.0mm}
\begin{tabular}{ccccc}
\toprule
Perceptual module   & Dynamic module  & MSE$\times 10^{-3}$(↓)   & LIPIS $\times 10^{-2}$(↓)\\
\midrule
\multirow{3}{*}{\begin{tabular}[c]{@{}c@{}}\\ Only position\end{tabular}} 

& Struct-VRNN   & 13.71      & 12.73     \\
 & V-CDN        & 10.76      & 9.27     \\
  & Grid keypoint   & 13.58            & 11.05     \\
 & Ours         & \textbf{8.73}     & \textbf{8.52}    \\
 \bottomrule
\end{tabular}
\end{table}

Experiments are conducted to verify the effectiveness of key components on the performance of the proposed model. The quantitative results of ablation experiments are reported in Table \ref{ablation_results}. We use the degraded versions of our model as the baseline model, where context-aware aggregator $\mathcal{A}$ is discarded and all Interaction networks in dynamic predictor $\mathcal{P}$ are replaced by MLPs (Multi-layer Perceptrons) with the same hidden layers. As Table \ref{ablation_results} shows, The aggregator $\mathcal{A}$ and IN help enhance the predictive ability, especially in Balls scenario. This is because that MLPs only consider the state of each object itself but discard interactions between them when forecasting future states. In Blocktower scenario, objects rarely interact with others and the environment, so simple object representation and dynamic modules are sufficient for the prediction task. Our model takes into account both object-object and object-environment interaction to better learn the dynamics behind physical systems, thus making more accurate predictions about the future states of objects. Besides, we find that skip connection in ResIN block plays a key role in the stable convergence of the proposed model during training. 
\begin{table}[htbp]
\renewcommand{\arraystretch}{1.3}
\caption{An ablation study on the context-aware aggregator $\mathcal{A}$ and Interaction network. We report  MSE of the prediction of object states and LPIPS of generated future frames for $t \in [0,\Delta T]$. }
\centering
\label{ablation_results}
\setlength{\tabcolsep}{2.6mm}
\begin{tabular}{c|cc|cc}
\toprule
Scenario & $\mathcal{A}$  & \multicolumn{1}{c|}{IN} & MSE$\times 10^{-3}$(↓)   & LPIPS$\times 10^{-2}$(↓)  \\
\midrule
\multirow{3}{*}{Blocktower} & $\times$ & $\times$  & 2.23 & 2.73  \\
& $\times$    & \checkmark   & 2.08 & 2.60  \\
& \checkmark  & \checkmark    & 2.01 & 2.50  \\
\midrule
\multirow{3}{*}{Balls}  & $\times$ & $\times$  & 3.34 & 6.15  \\
 & $\times$   & \checkmark   & 0.65 & 2.49  \\
 & \checkmark  & \checkmark  & 0.59 & 1.90\\
 \bottomrule
\end{tabular}
\end{table}

\section{Conclusion}
Inspired by recent works in compositionality and dynamics learning, we present an unsupervised framework for object-centric visual prediction, which disentangles static spatial information and dynamic variation from video representation and learns dynamics in low-dimension state space. Our model is composed of two key components. The perceptual module decomposes visual scenes into spatial feature maps and multiple physical meaningful object representations. Then, the dynamic module aggregates contextual information, infers physical interaction between objects and performs object-centric predictions. The context-aware aggregator and interaction-aware predictor integrate environment-object and object-object interaction mechanisms in physical systems, encouraging our model to learn more general physical dynamics. Future frames are generated by fusing spatial features and predicted states of objects. Experimental results demonstrate that our model outperforms other prediction methods and high-fidelity predictions are produced due to the learning of physical dynamics.

% or


% \appendices

% % you can choose not to have a title for an appendix
% % if you want by leaving the argument blank
% \section{}

% \section*{Acknowledgment}


\ifCLASSOPTIONcaptionsoff
  \newpage
\fi

\bibliographystyle{ieeetr}
% \bibliography{causal_and_physical_reasoning.bib}
\bibliography{ref.bib}
% \bibliographystyle{IEEEtran}
% \bibliography{IEEEabrv,causal_and_physical_reasoning}

% \begin{IEEEbiography}
% [{\includegraphics[width=1in,height=1.25in,clip,keepaspectratio]{pics/pic.jpg}}] 
% Huilin Xu (Student Member, IEEE) was born in Zhejiang Province, China, in 1999. She received the B.S. degree in communication science and engineering from Fudan University, Shanghai, China, in 2021. She is currently working toward the Ph.D. with the Key Laboratory of Information Science of Electromagnetic Waves (MoE), Fudan University. Her research interests include visual reasoning, causal inference and deep learning. 

% \end{IEEEbiography}
\end{document}